\documentclass[lettersize,journal]{IEEEtran}
\usepackage{amsmath,amsfonts}
\usepackage{amssymb}
\usepackage{algorithm}
\usepackage{algpseudocode}
\usepackage{array}
\usepackage[font=normalsize]{subcaption}
\usepackage{textcomp}
\usepackage{stfloats}
\usepackage{url}
\usepackage{verbatim}
\usepackage{graphicx}
\usepackage{cite}
\usepackage{multirow}
\usepackage{xr}
\usepackage{hyperref}
\usepackage{gensymb}
\begin{document}

\title{Disassociating Performance from Compositional Feature Learning}

\author{George Dimitriadis, Spyridon Samothrakis
\thanks{George Dimitriadis is with the Computer Science and Electronic Engineering Department of the University of Essex. email: gd25222@essex.ac.uk}
\thanks{Spyridon Samothrakis is with the Computer Science and Electronic Engineering Department of the University of Essex. email: ssamot@essex.ac.uk}
\thanks{The article has code at \href{https://github.com/georgedimitriadis/latent\_ood\_in\_world_models/tree/release\_v2}{https://github.com/georgedimitriadis/\newline latent\_ood\_in\_world\_models/tree/release\_v2}}}

\markboth{IEEE TRANSACTIONS ON COGNITIVE AND DEVELOPMENTAL SYSTEMS}%
{Shell \MakeLowercase{\textit{et al.}}: A Sample Article Using IEEEtran.cls for IEEE Journals}

\maketitle

\begin{abstract}
Out-of-distribution (OOD) generalisation through composition requires a system to discover invariant properties from input--output associations and transfer them to novel inputs and unseen tasks. We argue that confirming compositional learning requires more than OOD evaluation alone: one must also verify that the learned features are genuinely compositional and that the system encodes their compositional rules. We demonstrate this through two tasks with clearly defined OOD metrics, generated via composable high-level abstractions, on which three standard architectures (MLP, CNN, Transformer) and an object-centric, slot-based architecture fail to generalise OOD. We pair these tasks with two novel attention-based architectures featuring an interpretable final hidden layer designed to expose whether compositional representations emerge. One architecture carries an engineered inductive bias that enables near-perfect OOD performance on one task. Our results show that even with appropriate biases and near-perfect OOD accuracy, a model can fail to learn the compositional feature structures necessary for systematic generalisation. The interpretable layer reveals that successful OOD performance is driven by task-specific biases rather than the discovery of reusable compositional primitives. These findings indicate that OOD benchmarks alone are insufficient for evaluating compositionality in neural networks.
\end{abstract}


\section{Introduction}
\label{introduction}

A prominent goal of AI research is to develop machines with human-level cognitive capabilities \cite{collinsBuildingMachinesThat2024}. Despite extraordinary progress over the last few decades, human and animal cognition still exhibits attributes that elude the field's state-of-the-art models. Modern AI lacks, for example, the ability of humans and animals to accurately metacognise \cite{footeMetacognitionRat2007}, to transfer knowledge to new problems with vastly different input and output distributions \cite{lazaro-gredillaImitationZeroshotTask2019}, and to generate causal structures from very few examples \cite{gopnikTheoryCausalLearning2004}. It is widely held \cite{luettgauDecomposingDynamicalSubprocesses2024, szaboCaseCompositionality2012, gontierCombinatorialityCompositionalityEveryday2024, dehaeneSymbolsMentalPrograms2022, goyal_inductive_2022} that a fundamental property underlying these capabilities is \textit{compositionality}: the ability to extract higher-order representations (concepts) and high-level compositional functions (rules) from input data, and to recombine them to solve out-of-distribution (OOD) problems and reason about counterfactuals.

As the field transitioned from innately compositional symbolic methods to deep learning \cite{russinFregeChatGPTCompositionality2024}, the question of how to reintroduce compositionality without sacrificing the strengths of neural methods has driven a rich body of work. This work spans neurosymbolic AI \cite{garcezNeurosymbolicAI3rd2023}, probabilistic program induction \cite{ellisDreamCoderBootstrappingInductive2021a, collinsBuildingMachinesThat2024}, modular deep neural networks \cite{goyal_inductive_2022}, disentangled representation learning \cite{higginsBetaVAELearningBasic2017}, object-centric learning \cite{wuNeuralLanguageThought2023, kipfContrastiveLearningStructured2020}, and chain-of-thought reasoning \cite{venkatramanAmortizingIntractableInference2024}. Nevertheless, productivity and systematicity---two core aspects of compositionality identified by Hupkes et al.\ \cite{hupkesCompositionalityDecomposedHow2020}---remain largely unattained by current architectures, lending renewed force to Fodor and Pylyshyn's classical compositionality challenge \cite{fodorConnectionismCognitiveArchitecture1988, russinFregeChatGPTCompositionality2024}.

In this work, we focus on systematicity: the recombination of known parts and rules to understand novel compositions \cite{hupkesCompositionalityDecomposedHow2020}, viewed through the lens of functional compositionality as formalised by van Gelder \cite{vangelderCompositionalityConnectionistVariation1990a}. We argue that a model exhibiting genuine systematicity will display two characteristics. First, it will demonstrate OOD generalisation across a range of tasks, not only the single task on which it was designed to be evaluated. Second, its latent representation space will exhibit a structure akin to symbolic coding rather than statistical learning: a small (but not overly small) number of features encoding addressable, composition-amenable symbols, rather than a large number of features interacting through weak correlations (see Fig.~\ref{fig:concept}).

Building on theoretical work by Thorpe \cite{thorpeLocalVsDistributed1989} and the recent complexity-based framework of Elmoznino et al.\ \cite{elmozninoComplexityBasedTheoryCompositionality2025}, we argue that many results in the literature on compositional learning (e.g., \cite{lakeHumanlikeSystematicGeneralization2023}), although seemingly OOD, do not give rise to such latent features. Their success can be attributed to the model learning the distribution of the task or meta-task, a feat that does not require compositionality. We support this argument by demonstrating how modern architectures can fail to discover the latent feature spaces required for compositional OOD, and how architectures that exhibit task-specific OOD generalisation can do so without employing compositional strategies. We are aware of the large number of specialised architectures that solve tasks through compositional representation discovery (see Section~\ref{ai_approaches} for a non-comprehensive list). However, in their overwhelming majority, such networks are engineered with appropriate biases to achieve compositionality on their target tasks. We contend that in such efforts the hard problem is solved by human engineers who embed the necessary biases into their networks, or through careful design of the task space that gives models only the appearance of having learned compositional latent structure supporting OOD generalisation. Our thesis targets networks that lack explicit compositionality biases and are presumed capable of discovering compositional representations on arbitrary tasks---much like human and animal brains---simply because they exhibit OOD generalisation on a small number of benchmarks.

To make this case, we construct two data sets arising from two world models, based on the input--output design of the Abstract and Reasoning Corpus (ARC) competition \cite{cholletMeasureIntelligence2019, cholletFcholletARCAGI2025}. Our tasks are built on parametrised objects with fully defined composition rules, where objects and transformations are separated at the level of the task input, providing the learning algorithm with a disentangled input representation. A visual rather than a token-sequence paradigm (such as the SCAN benchmark \cite{zerrougBenchmarkCompositionalVisual2022a}) was chosen for three reasons. First, newly emerging differences between human and AI visual task solutions \cite{depewegSolvingBongardProblems2024} suggest that AI algorithms, despite high accuracy scores, do not follow compositional strategies. Second, we sought to avoid certain assumptions about syntactic rules that are often implicit in sequence tasks, as discussed by van Gelder \cite{vangelderCompositionalityConnectionistVariation1990a}. Third, work with similar goals and conclusions has already been conducted on sequence tasks by Dasgupta et al.\ \cite{dasguptaAnalyzingMachineLearnedRepresentations2020}, who found that models trained on simple sequence tasks learn non-compositional heuristics rather than compositional representations.

Our world models follow the structure presented in \cite{okawaCompositionalAbilitiesEmerge2023}, which yields a clear definition of OOD through a distance measure between test and training sets. We extend this structure by adding two variables, one semantic (object size) and one geometric (object initial position). These data sets are designed to be shallow in their compositionality depth (detect a single object, find a single rule, apply the rule to the object), a simplicity that affords an intuitive understanding of what compositionality and OOD generalisation should look like. Despite this apparent simplicity, our tasks demand not just a semantic compositional capability (e.g., change colour then change shape) but one that survives geometric changes (e.g., moving or rotating an object between positions never seen during training). In this respect, they add the rule in the compositional chain, demanding from the benchmarked algorithms to compose not only objects and object attributes with each other but also object transformations (actions) with objects. Finally, they allow a discrete measure of whether an algorithm correctly encodes an object and a transformation rule (much like the ARC tasks from which they are derived). Because of this discrete nature, our world models admit as a test metric not a continuous, fuzzy measure (such as mean squared error), where meaningful errors can be masked by averaging over pixels, but a discrete one (correct image or not), where only a true conceptualisation of the objects and their one-step transformation rules can yield a correct solution.

We evaluate these data sets on three groups of architectures. The first group comprises three commonly used architectures---an MLP, a CNN, and a Transformer---none of which is known for solving problems compositionally. These serve as baselines that compositional models should be expected to surpass. The second group consists of a single architecture with an object-centric inductive bias, the Object-Centric Learning (OCL) model of \cite{dedhiaImPromptuIncontext2023}, designed to detect objects in images and compose object-manipulation rules over them. OCL is a slot-based architecture currently considered state-of-the-art (SOTA) among object-centric, compositionally based algorithms. The third group consists of two architectures of our own design. These share three key properties: they achieve substantially better OOD performance on our data sets than both the standard architectures and the object-centric one; they possess an interpretable final latent layer that can be coherently visualised, enabling unambiguous detection of whether compositional representations have been learned; and the two architectures differ only in that one carries an additional inductive bias that makes one of the two data sets more OOD-learnable, thereby clearly illustrating how an engineered rather than learned bias can be the sole driver of apparent OOD behaviour while degrading performance on tasks whose solution does not align with it.

We propose that evaluating algorithms as ``compositional'' based solely on their results on specific benchmarks can often be misleading, especially when the task/metric combination can obscure the true nature of an algorithm's learned representations, even when those results indicate a high degree of OOD learning. For AI to achieve compositionality, a more nuanced evaluation is required---one that characterises new algorithms both with a diverse set of OOD benchmarks and with methods that directly probe the structure of their latent feature spaces.

The remainder of this paper is organised as follows. Section~\ref{background} reviews the concept of compositionality and relevant prior work. Section~\ref{methods} describes the methodologies used for generating the data sets and the models evaluated. Section~\ref{results} presents our results, demonstrating a clear case in which models that achieve high OOD accuracy (given appropriate biases) still fail to develop latent features suitable for compositional generalisation over feature-invariant tasks. Section~\ref{discussion} discusses the main limitations and avenues for future work.

\section{Background}
\label{background}

\subsection{Compositionality: An Overview}
\label{overview}

Compositionality was originally introduced as a key aspect of human language \cite{szabo_compositionality_2004, szaboCaseCompositionality2012, werningOxfordHandbookCompositionality2012} and as a common property of language and the way minds understand the world \cite{fodor_language_1980}. In linguistics, compositionality is defined as ``The meaning of a whole is a function of the meanings of the parts and of the way they are syntactically combined'' \cite{parteeLexicalSemanticsCompositionality1995}. In the behavioural sciences, compositionality refers to ``hierarchical nesting of parts into larger wholes with new individuality, identity, or meaning---or inversely, partitioning of such wholes into smaller parts with independence or individual meaning of their own'' \cite{gontierCombinatorialityCompositionalityEveryday2024}. A substantial body of work has further shown that not only humans but also animals approach the decomposition of sensory input and the composition of actions in a compositional fashion \cite{zentallConceptLearningAnimals2008, battagliaStructuredCognitionNeural2012, gontierCombinatorialityCompositionalityEveryday2024}.

The observation that nature appears compositional in structure motivates the assumption that cognition achieves two goals. The first is to extract from input data higher-order representations (concepts) as well as high-level compositional functions (rules) that operate on both concepts and on themselves (see \cite{vangelderCompositionalityConnectionistVariation1990a} for a definition of composable concepts and rules; see also \cite{ohlssonAbstractionAcquisitionComplex1997} for a counterposition on humans' ability to extract concepts and rules from data). The second is to use those abstractions to solve OOD problems and generate possible solutions to counterfactuals. In humans, such abstractions bind together and render accessible large swathes of sensory and action spaces, most of which are inexperienced or even physically impossible \cite{ohlssonAbstractionAcquisitionComplex1997, tenenbaumHowGrowMind2011a}.

\subsection{Modern Approaches to Compositionality in Machine Learning}
\label{ai_approaches}

As the field of AI transitioned from symbolic methods to connectionist and deep learning approaches \cite{russinFregeChatGPTCompositionality2024}, the argument for the importance of compositionality remained strong \cite{fodorConnectionismCognitiveArchitecture1988}. Van Gelder placed the concept of compositionality within connectionist approaches on a firm theoretical basis \cite{vangelderCompositionalityConnectionistVariation1990a}. From that point, a proliferation of methodologies has followed, including neurosymbolic AI \cite{garcezNeurosymbolicAI3rd2023}, probabilistic program induction \cite{ellisDreamCoderBootstrappingInductive2021a, collinsBuildingMachinesThat2024}, modular deep neural networks \cite{goyal_inductive_2022}, disentangled representation learning \cite{higginsBetaVAELearningBasic2017}, object-centric learning \cite{wuNeuralLanguageThought2023,kipfContrastiveLearningStructured2020, dedhiaImPromptuIncontext2023}, and chain-of-thought reasoning \cite{venkatramanAmortizingIntractableInference2024}. For reviews, see also \cite{linSurveyCompositionalGeneralization2023, sinhaSurveyCompositionalLearning2024}, and \cite{russinFregeChatGPTCompositionality2024}.

Hupkes et al.\ \cite{hupkesCompositionalityDecomposedHow2020} addressed the application of the traditional linguistic notion of compositionality within a modern machine learning context. They decomposed the concept into five underlying notions: systematicity, productivity, substitutivity, localism, and overgeneralisation. They then constructed theoretically grounded compositionality benchmarks for three commonly used neural network architectures: Long Short-Term Memory networks (LSTM) \cite{hochreiterLongShortTermMemory1997}, Convolutional Networks (CNN) \cite{NIPS1989_53c3bce6}, and Transformers \cite{vaswaniAttentionAllYou2023}. Their findings indicated that productivity and systematicity remain unattainable by current models, with Transformers performing better than the other two architectures. These results have renewed attention to Fodor and Pylyshyn's argument \cite{fodorConnectionismCognitiveArchitecture1988} regarding the compositional deficiency of connectionist architectures, also known as the compositionality challenge of neural networks \cite{russinFregeChatGPTCompositionality2024}. One of the most recent efforts towards addressing this challenge was the meta-learning-based approach of Lake and Baroni \cite{lakeHumanlikeSystematicGeneralization2023}, who showed that on a series of linguistic tasks, an attention-based model could learn novel, unseen tasks (automatically generated by a pre-specified process) through the use of only a few in-context examples. In the field of image recognition and processing, object-centric, slot-attention-based algorithms such as the OCL architecture of Dedhia et al.\ \cite{dedhiaImPromptuIncontext2023} (a combination of the SLATE algorithm \cite{singhIlliterateDALLELearns2022} and an attention-based analogy-discovery process) are purported to appropriately encode object-centric structures and composable object transformations.

\begin{figure*}[h!]
    \centering
     \subfloat[Classical machine learning]{%
        \includegraphics[trim=0 0 0 2.3cm, clip, width=0.33\textwidth]{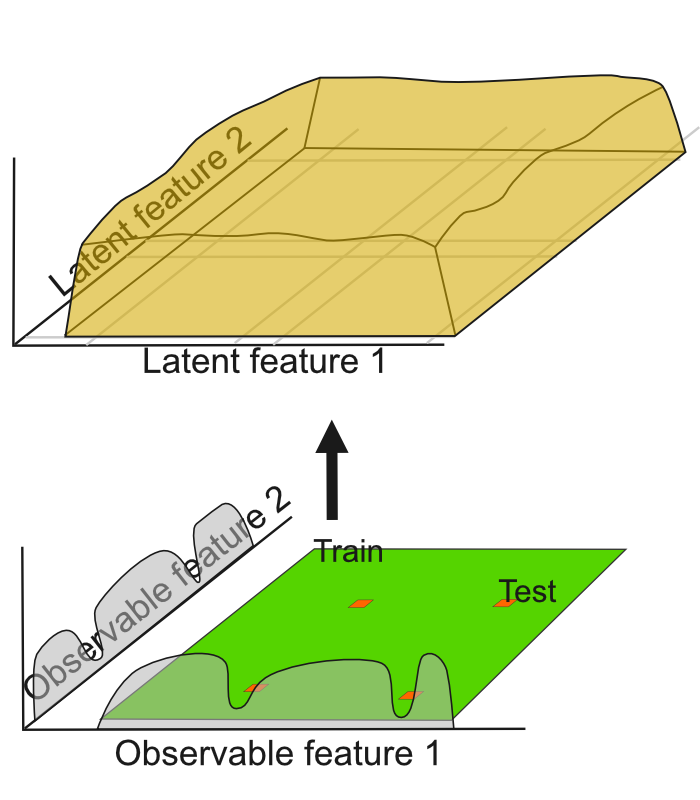}
    }
    \subfloat[Meta learning]{%
        \includegraphics[trim=0 0 0 2.3cm, clip, width=0.33\textwidth]{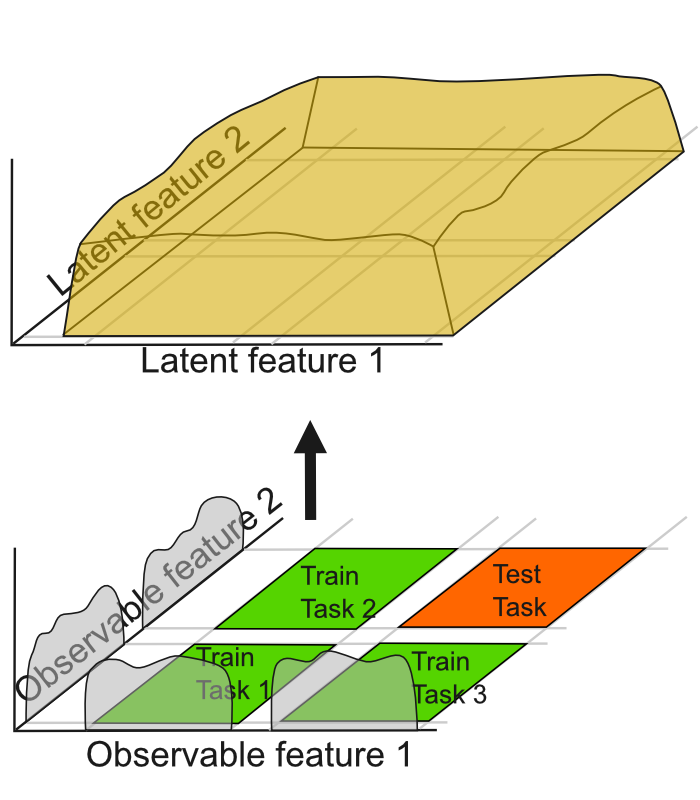}
    }
   \subfloat[OOD learning]{%
        \includegraphics[trim=0 0 0 2.3cm, clip, width=0.33\textwidth]{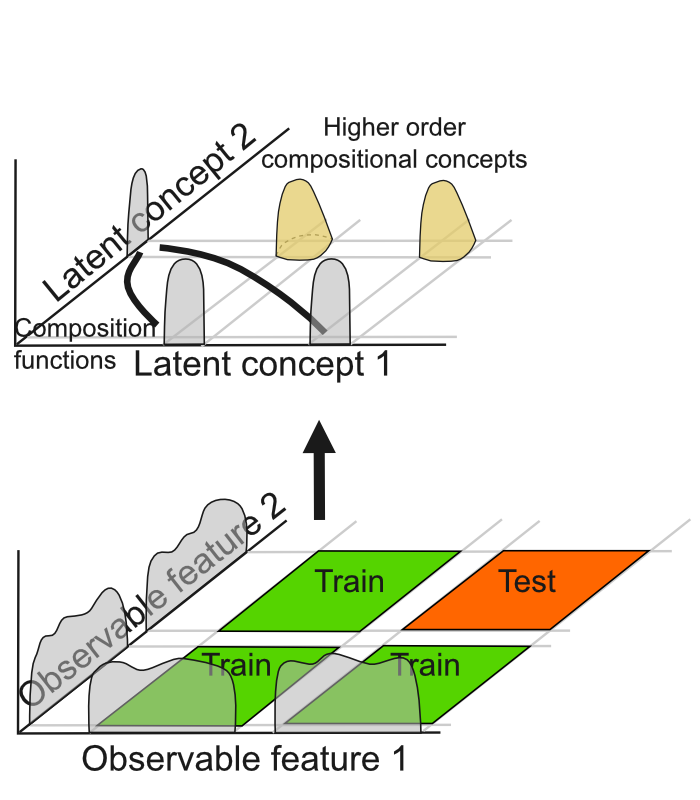}
    }
    \caption{A conceptual understanding of the relationship between classical machine learning algorithms, modern meta learning ones and a working out of distribution algorithm based on compositionality. We argue that current ML frameworks, both from classical ML and from meta-learning approaches generate latent features in spaces (yellow regions) with no real compositional capability. These representations are spread throughout the latent space and operate on the base of a large number of small correlations between subregions. A true compositional architecture would be able to discover latent features whose distribution over their spaces would appear much more multimodal and discreet (concepts), allowing them to combine those concepts into higher order compositions without modes of failure arising from statistical mixing of broad distributions.}
    \label{fig:concept}
\end{figure*}

\subsection{Systematicity in Representation}
\label{syst_in_repr}

In this work, we focus exclusively on systematicity as defined by Hupkes et al.: ``This ability concerns the recombination of known parts and rules: anyone who understands a number of complex expressions also understands other complex expressions that can be built up from the constituents and syntactical rules employed in familiar expressions.'' from the viewpoint of functional compositionality as proposed by van Gelder. Functional compositionality does not assume concatenation-based syntactic rules (that compose tokens into larger expressions) and is a form of compositionality more consistent with connectionist approaches \cite{vangelderCompositionalityConnectionistVariation1990a}.

We argue that a model exhibiting systematicity will display two characteristics. First, it will demonstrate OOD generalisation across a range of tasks (as do animals and humans), and not only on the single task on which it was designed to be evaluated. Second, its latent representation space (if appropriately probed) will exhibit a structure resembling symbolic coding rather than statistical learning. Specifically, there will be a small (but not overly small) number of features encoding addressable and composition-amenable symbols, rather than a large number of features interacting through an even larger number of weak correlations (see Fig.~\ref{fig:concept} for a schematic illustration). This characteristic derives from the work of Thorpe \cite{thorpeLocalVsDistributed1989} (see also the analysis by Olah \cite{olahDistributedRepresentationsComposition2023}) and, more recently, Elmoznino et al.\ \cite{elmozninoComplexityBasedTheoryCompositionality2025}. Thorpe's work establishes that composable representations require a certain degree of compressibility between one-to-one mapping and maximally compressed coding. Elmoznino et al.\ place this intuition on a theoretical basis, arguing that a compositional representation is a semantic mapping from a set of sentences composed of symbolic features (tokens) to a set of continuous vectors, with the tokens, their syntax, and the semantic map chosen such that the representation is maximally compressed in the Kolmogorov complexity sense. Their work highlights the difficulty of discovering such a language (tokens and their syntactic rules) and then representing it in a model's continuous latent features.

We argue that many results in the literature on compositional learning (e.g., \cite{lakeHumanlikeSystematicGeneralization2023}), although seemingly OOD, do not give rise to the required latent features. Their success on the evaluated tasks is simply the result of the model learning the distribution of the task or meta-task (i.e., the set of all outputs the data-generating process can produce), a feat that does not require compositionality.

\section{Methods}
\label{methods}

\subsection{World Models and Data Sets}
\label{models_data}
To validate our thesis, we construct two data sets derived from two distinct world models. These share identical input and output observation spaces but differ in their action spaces and the corresponding distributions over observables. Both world models operate on images structurally analogous to those in the ARC Competition \cite{cholletFcholletARCAGI2025}. Each image consists of a 32\,$\times$\,32 pixel grid where each pixel takes one of four colour values. White pixels denote unused space, black pixels define an $n \times m$ background canvas ($n, m \leq 32$), and pixels of the remaining two colours (red and blue) represent objects placed on this canvas. Each input--output pair contains a single coloured object, consistent across both images, defined by a set of coloured pixels. Objects may extend partially beyond the canvas boundary, in which case only the visible portion is rendered.

\begin{figure*}[ht]
\centering
     \subfloat[Translate]{%
        \includegraphics[width=0.45\textwidth]{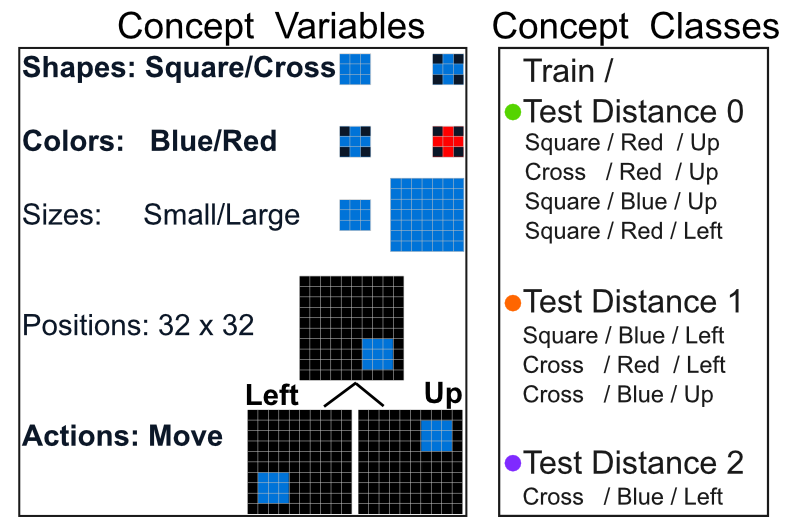}
    }
    \subfloat[Rotate]{%
        \includegraphics[clip, width=0.45\textwidth]{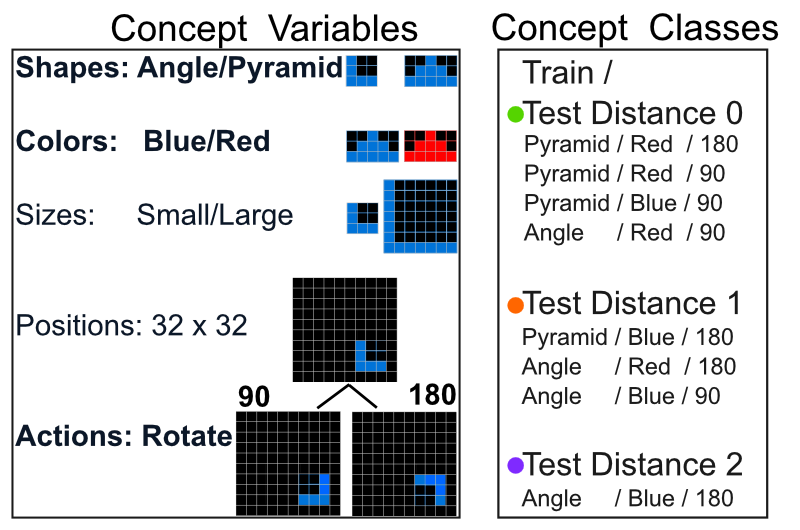}
    }\hfill
    \subfloat[Concept graph]{%
        \includegraphics[width=0.45\textwidth]{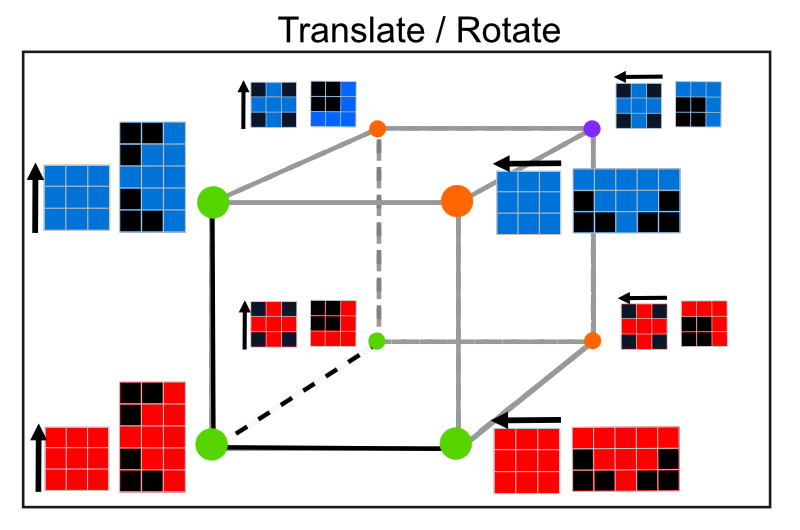}
    }
    \subfloat[Common network architecture]{%
        \includegraphics[clip, width=0.45\textwidth]{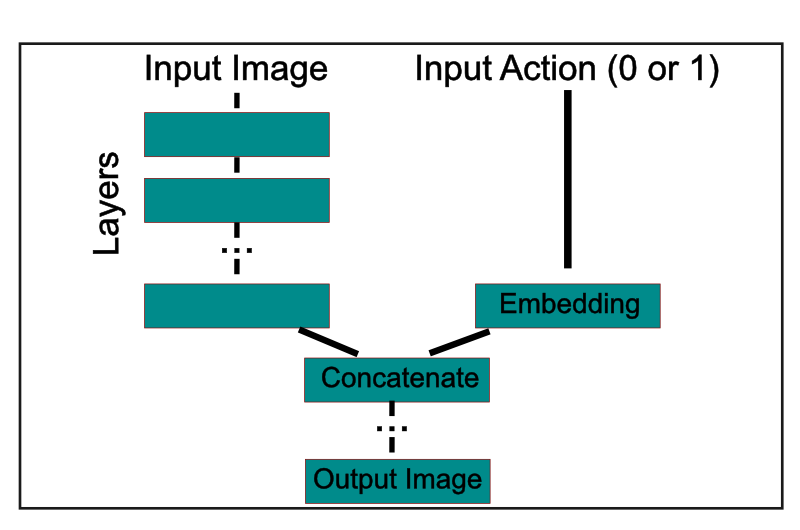}
    }\hfill
\caption{Overview of our methodology. (a) The design of the Translate world model with its derived data sets. (b) The design of the Rotate world model with its derived data sets. (c) The Concept Graphs for both world models showing which combinations of CVs belong to which data set (also shown in a and b in the Concept Classes boxes) and the distance between each combination on the unit cube. So a Train or a Test Distance 0 sample would be derived from combining the values in the Train / Test Distance 0 column in b), shown diagrammatically as the green dots of the cube in c). For example a possible Train sample would be a blue square \textbf{moving Up} or a red pyramid \textbf{rotating 90} (top left front green corner of the Concept Graph cube. A Distance 2 sample could only be a blue cross \textbf{moving Left}  or a blue angle \textbf{rotating 180}  (seen in the top right back purple corner of the Concepts graph) (d) The common architecture for the algorithms in this work that take as input the images and the language bit denoting the transformation rule. The architecture of the OCL algorithm was identical as the one used in its presenting publication. The type of network defined the left hand path types of layers and the embedding size of the right hand path.}
\label{fig:data}
\end{figure*}

The design of these two world models follows the compositional structure proposed in \cite{okawaCompositionalAbilitiesEmerge2023}. Each model is parameterised by five independent concept variables (CVs) that govern the appearance and transformation of objects within the images. Four of these are binary-valued and one takes values in $\mathbb{N}^{32 \times 32}$. Across both models, the CVs are: \emph{Shape}, \emph{Colour}, \emph{Size}, \emph{Position} on the canvas, and \emph{Action}, which specifies the transformation applied to the object to produce the target output (see Fig.~\ref{fig:data}A and B for the domain of each CV).

Each world model is named according to its \emph{Action} CV. The \textbf{Translate} model defines actions \textbf{Move Left} and \textbf{Move Up} (each by 6 pixels), while the \textbf{Rotate} model defines actions \textbf{Rotate 90} and \textbf{Rotate 180}. For each world model, a training set and multiple held-out test sets are generated to evaluate OOD generalisation. All possible combinations of three CVs (\emph{Shape}, \emph{Colour}, \emph{Action}) are enumerated, yielding a concept graph of 8 independent combinations (see Fig.~\ref{fig:data}C). These combinations are partitioned into three concept classes. The class containing four combinations is used to construct the training set and an in-distribution test set (denoted Test Distance~0). The remaining four combinations form two additional OOD test sets: one comprising combinations at graph distance~1 from the training distribution (Test Distance~1), and one comprising the single combination at graph distance~2 (Test Distance~2). The remaining two CVs (\emph{Position} and \emph{Size}), as well as the canvas dimensions, are sampled uniformly at random across all data sets, subject to a minimum canvas size of 10 pixels in each dimension.

Each sample consists of an input tuple paired with a target output image. The input tuple comprises the source image and a binary action indicator specifying the transformation to be applied. As a concrete example from the Translate world model: the input image might depict a large (\emph{Size}) 7\,$\times$\,7 pixel blue (\emph{Colour}) square (\emph{Shape}) with its bottom-left pixel at grid position (\emph{Position}) (6,\,1) on a 10\,$\times$\,10 black canvas. The action indicator (\emph{Action}) is 0 for \textbf{Move Left} or 1 for \textbf{Move Up}. The corresponding target image contains the same canvas and object, translated by 6 pixels in the direction specified by the action indicator.

\subsection{Algorithms}
\label{algos}
We evaluate the OOD generalisation capabilities of six architectures on the above data sets: an MLP, a CNN, a Transformer, a slot-attention model with an object-centric inductive bias, and two bespoke attention-based networks designed to achieve strong OOD performance on one or both world models.

The five architectures other than OCL share a common dual-pathway input structure that merges into a single output pathway (see Fig.~\ref{fig:data}D). The first pathway processes the input image, with the specific layer types determined by the architecture. The second pathway receives the binary action indicator, which is projected into a learned embedding vector whose dimensionality is architecture-dependent. The image representation and action embedding are concatenated and passed through subsequent architecture-specific layers to produce the output. The OCL model is instead designed to discover the rule from a number of extra image pairs provided as examples (referred to in the original work as analogies); here we used two analogy pairs per sample, drawn from the same distribution defining the class of the sample (Train, or Distance~0, Distance~1, or Distance~2, and Translate or Rotate). To ensure a fair comparison, all models are matched as closely as possible in total parameter count: the Transformer has 4.6M parameters, the CNN 4.5M, the MLP 6.2M, and both axial pointer networks 4.4M. OCL also has a comparable number of parameters (5.5M) (see Table~\ref{table:num_of_params} for details).

The hyperparameters of the OCL model were set to match the configuration of the original publication \cite{dedhiaImPromptuIncontext2023} as closely as possible, using the code from its GitHub repository directly. We used two analogy image pairs per sample (a 2-shot setting), three slots, and a vocabulary size of 128 in the SLATE algorithm, with 30,000 steps for the SLATE pre-training stage and 150 epochs (80,000 training samples and 20,000 validation samples) for OCL training. The remaining hyperparameters follow \cite{dedhiaImPromptuIncontext2023} and \cite{singhIlliterateDALLELearns2022}. The choice of three slots reflects the three ``objects'' the algorithm should be able to detect and separate: the black background, the white surround, and the coloured object.

The two bespoke architectures incorporate gated MLPs (gMLP) \cite{liuPayAttentionMLPs2021}, axial attention \cite{hoAxialAttentionMultidimensional2019}, and pointer networks \cite{vinyalsPointerNetworks2015}. The pointer network layer implements an attention mechanism that learns a copy operation from input to output pixels, effectively computing a mapping from each input pixel index to a corresponding output pixel index. This architectural choice yields an interpretable mapping that can be directly inspected to analyse how each model transforms inputs across the different data sets (see Section~\ref{results} and Fig.~5). The axial attention mechanism, integrated with the pointer attention layer, factorises the 2D index prediction into independent computations along each spatial axis.

Specifically, in the axial pointer network (APN), the $x$ and $y$ components of each pixel's destination index are predicted independently. We introduce a variant, the axial pointer linear network (APLN), which first predicts the target column index and subsequently predicts the row within that column. In the algorithms panel we show the inference and training loop, common to the two algorithms (Algorithm~\ref{alg:apn-point} and Algorithm~\ref{alg:apn-train}) as well as their diverging part in the training (Algorithm~\ref{alg:apn-forward}). The difference is in the pointer factorisation, i.e. the shape and meaning of the logits the control MLP emits: the APN emits an independent axial pointer per target pixel (\verb|[H,W,H]| + \verb|[H,W,W]|), while the APLN emits a single global row-mixing matrix and a single column-mixing matrix (\verb|[H,H]| + \verb|[W,W]|) applied to the whole canvas. The lines variant is therefore a hard architectural prior that the task is a whole-line rearrangement, bought with a 32× smaller output head. This decomposition introduces an inductive bias favouring column- and row-wise copying operations consistent with translational transformations, and was designed to facilitate OOD generalisation on the Translate data sets while potentially hindering performance on the Rotate data sets. Our experimental results confirm this intended behaviour. Finally, we note that the gMLP layers play a critical role in producing well-calibrated logits for the attention mechanism; replacing them with standard MLP layers substantially degrades OOD generalisation performance (data not shown).


\begin{algorithm}[t]
\caption{Axial pointer network forward pass. The trunk is shared; the two
variants differ only in the size of the logit head and the granularity of
the copy operation.}
\label{alg:apn-forward}
\begin{algorithmic}[1]
\Require grid $X \in \{0,\dots,10\}^{H \times W}$, task index $z$, mode $\in \{\textsc{Full}, \textsc{Lines}\}$
\Ensure predicted grid $Y' \in [0,1]^{H \times W \times 11}$
\Statex
\Statex \textbf{Shared trunk}
\State $x \gets \mathrm{OneHot}(X, 11)$ \Comment{canvas to be copied from}
\State $\tilde{x} \gets x + \mathcal{N}(0, 0.05)$ \Comment{Gaussian noise, training only}
\State $p \gets \mathrm{PatchEncoder}_{256}\big(\mathrm{Patches}_{8\times 8}(\tilde{x})\big)$ \Comment{$16$ patches $+$ positional emb.}
\For{$g = 1, \dots, 10$}
    \State $p \gets \mathrm{gMLP}(p)$ \Comment{spatial-gating mixing}
\EndFor
\State $f \gets \mathrm{Flatten}(p)$
\State $e_z \gets \mathrm{Embedding}(z)$ \Comment{task embedding, $128$-d}
\State $c \gets \mathrm{MLP}_{\mathrm{res}}\big([\,f \,;\, e_z\,]\big)$ \Comment{control MLP; $32$ units (\textsc{Full}) / $256$ units (\textsc{Lines})}
\State $\ell \gets \mathrm{Dense}(c)$ \Comment{attention logits; \emph{size differs by variant}}
\Statex
\Statex \textbf{Divergent trunk - APN}
\If{mode $=$ \textsc{Full}} \Comment{\emph{pixel-level} pointers, $|\ell| = (H{+}W)HW = 65{,}536$}
    \State $L_h \gets \mathrm{reshape}\big(\ell_{[\,:HWH\,]}\big) \in \mathbb{R}^{H \times W \times H}$ \Comment{per target pixel: dist.\ over source rows}
    \State $L_w \gets \mathrm{reshape}\big(\ell_{[\,HWH:\,]}\big) \in \mathbb{R}^{H \times W \times W}$ \Comment{per target pixel: dist.\ over source cols}
    \State $P_h, P_w \gets \mathrm{Point}(L_h), \mathrm{Point}(L_w)$ \Comment{Alg.~\ref{alg:apn-point}}
    \State $T_{i,w,c} \gets \sum_h (P_h)_{i,w,h}\, x_{h,w,c}$ \Comment{gather along height}
    \State $Y'_{i,j,c} \gets \sum_w (P_w)_{i,j,w}\, T_{i,w,c}$ \Comment{then along width}
    \Statex \hspace{\algorithmicindent} $\triangleright$ each output pixel $(i,j)$ independently copies from its own chosen $(\text{row}, \text{col})$
\Statex
\Statex \textbf{Divergent trunk - APLN}
\ElsIf{mode $=$ \textsc{Lines}} \Comment{\emph{line-level} pointers, $|\ell| = H^2 + W^2 = 2{,}048$}
    \State $A_r \gets \mathrm{reshape}\big(\ell_{[\,:H^2\,]}\big) \in \mathbb{R}^{H \times H}$ \Comment{one global row $\to$ row map}
    \State $A_c \gets \mathrm{reshape}\big(\ell_{[\,H^2:\,]}\big) \in \mathbb{R}^{W \times W}$ \Comment{one global col $\to$ col map}
    \State $A_r, A_c \gets \mathrm{Point}(A_r), \mathrm{Point}(A_c)$ \Comment{Alg.~\ref{alg:apn-point}}
    \State $Y' \gets A_r\, x$ \Comment{move entire rows}
    \State $Y' \gets \big(A_c\, Y'^{\top}\big)^{\top}$ \Comment{then entire columns}
    \Statex \hspace{\algorithmicindent} $\triangleright$ all pixels of an output row come from one source row (likewise columns): only whole-line moves are expressible
\EndIf
\State \Return $Y'$
\end{algorithmic}
\end{algorithm}

\begin{algorithm}[t]
\caption{$\mathrm{Point}(L)$: soft pointer during training, hard pointer at evaluation.}
\label{alg:apn-point}
\begin{algorithmic}[1]
\Require logits $L$ (softmax over the last axis), temperature $\tau = 1$
\If{training}
    \State $u \sim \mathcal{U}(0,1)$; \quad $g \gets -\log(-\log u)$ \Comment{Gumbel noise}
    \State \Return $\mathrm{softmax}\big((L + g)/\tau\big)$ \Comment{soft, differentiable}
\Else
    \State \Return $\mathrm{OneHot}\big(\arg\max L\big)$ \Comment{hard, discrete copy}
\EndIf
\end{algorithmic}
\end{algorithm}

\begin{algorithm}[t]
\caption{Training and per-distance evaluation loop (identical for \textsc{Full} and \textsc{Lines}).}
\label{alg:apn-train}
\begin{algorithmic}[1]
\Require train set $(X, Z, Y)$; test sets $\{(X_d, Z_d, Y_d)\}_{d \in \{0,1,2\}}$ at compositional distance $d$
\State compile with AdamW ($\eta = 10^{-4}$, weight decay $0.004$), sparse categorical cross-entropy
\For{$\text{epoch} = 1, \dots, N$}
    \State fit one epoch on $(X, Z) \to Y$, batch size $100$
    \Statex \hspace{\algorithmicindent} $\triangleright$ forward uses noisy input and \emph{soft} Gumbel pointers: gradients flow through the copy op
    \For{$d \in \{0, 1, 2\}$}
        \State evaluate on $(X_d, Z_d, Y_d)$ and log score
        \Statex \hspace{2\algorithmicindent} $\triangleright$ evaluation uses \emph{hard} $\arg\max$ pointers: the network must commit to discrete copies
    \EndFor
    \If{$\text{epoch} \bmod n_{\mathrm{save}} = 0$} save model (and figures) \EndIf
\EndFor
\end{algorithmic}
\end{algorithm}

\begin{table}[h!]
\begin{tabular}{ |p{2.5cm}||p{2.5cm}|p{2.5cm}|}
 \hline
\textbf{Network} & \textbf{Total Parameters}  & \textbf{Trainable Parameters}\\
  \hline
MLP & 6,170,976 & 6,170,976  \\
CNN & 4,474,031 & 4,466,215 \\
Transformer & 4,663,040 & 4,663,040 \\
OCL & 5,525,891 & 5,525,891 \\
APN  & 4,481,088 & 4,480,384 \\
APLN & 4,446,112 & 4,440,480 \\
\hline
\end{tabular}
\\
\caption{Number of parameters for each of the six networks used.}
\label{table:num_of_params}
\end{table}

\section{Results}
\label{results}
\subsection{OOD Generalisation Is Achieved Only by the Appropriately Biased Architecture}
\label{ood_achieved}
\begin{figure*}[ht]
\centering
     \subfloat[Translate]{%
        \includegraphics[width=0.95\textwidth]{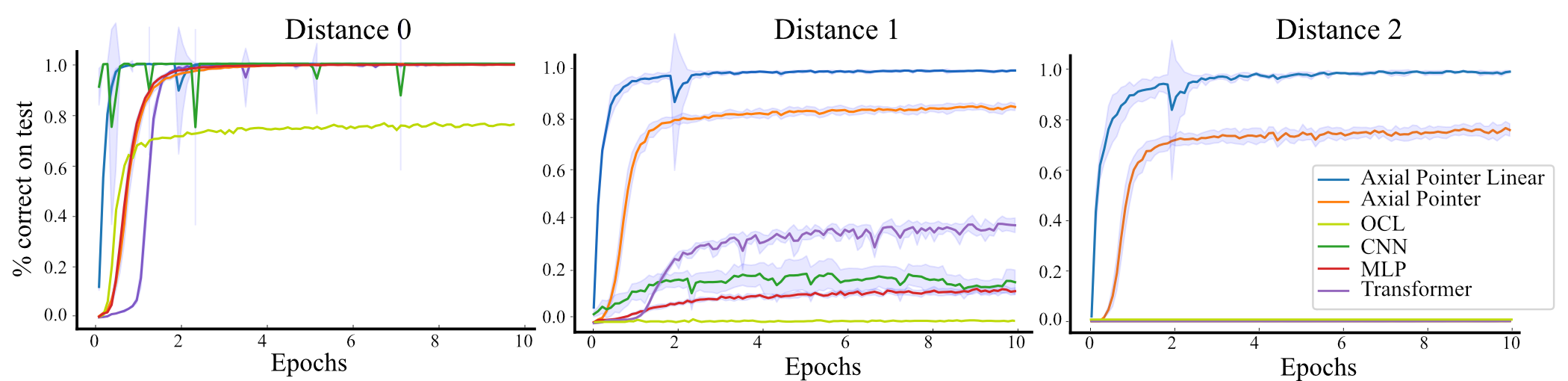}
    }\hfill
    \subfloat[Rotate]{%
        \includegraphics[width=0.95\textwidth]{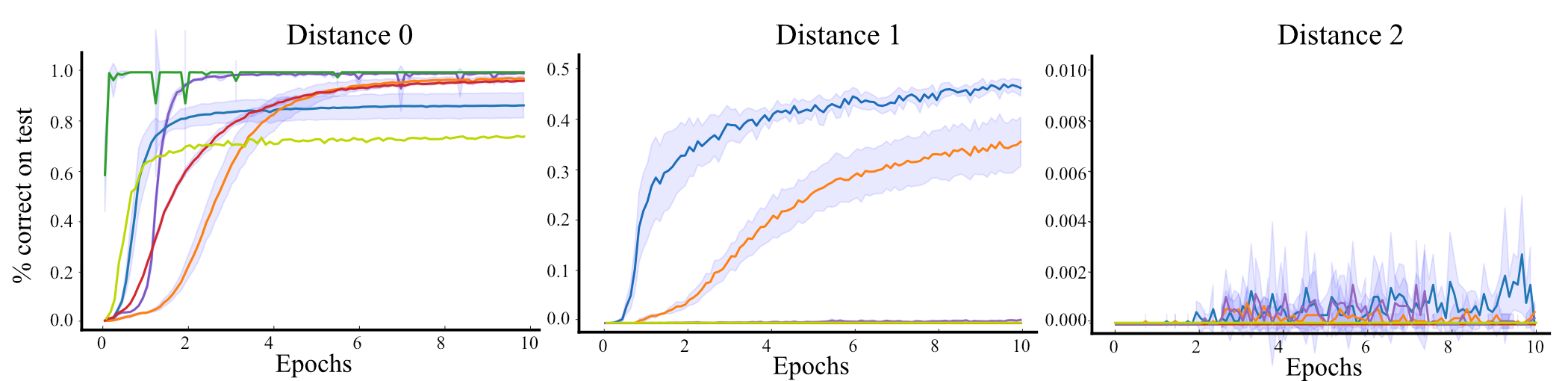}
    }\hfill
    \subfloat[Rotate for different APN and APLN sizes]{%
        \includegraphics[width=0.95\textwidth]{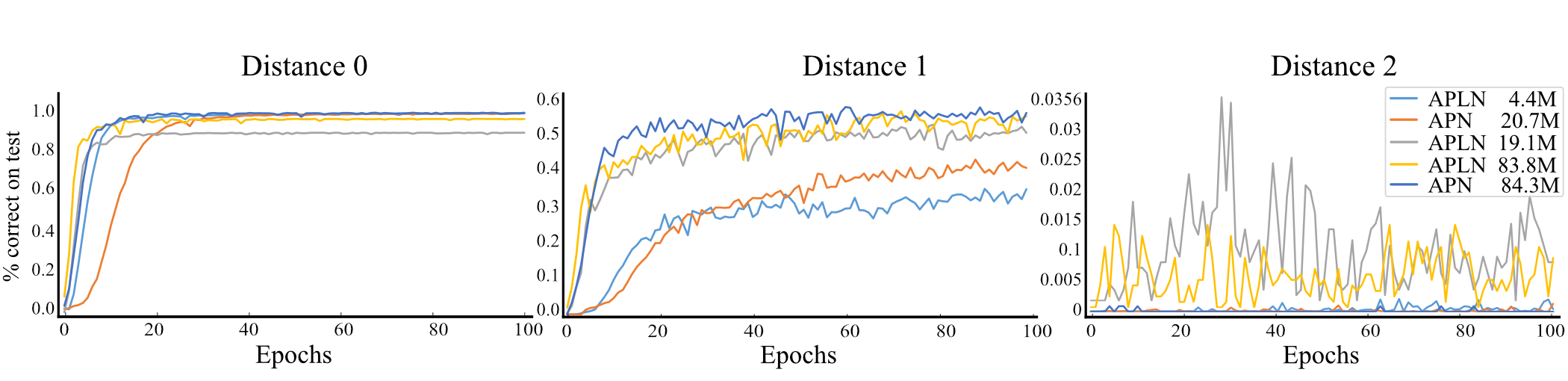}
    }
\caption{The results of all networks on the 6 test data sets arising from the two world models. The confidence intervals are generated from 10 runs with a 95\% confidence level. The results of all six networks on Translate (a) and Rotate (b) world models where all networks have their total number of parameters matched. (c) The results of the APN and APLN for the Rotate world model showing different versions of the networks with different number of total parameters.}
\label{fig:results}
\end{figure*}

Fig.~\ref{fig:results}(a) and (b) present the learning curves of all architectures across the six test sets (three from each world model). The reported metric is the percentage of fully correct output images produced by each network on each test set. As expected, all architectures perform well on the Distance~0 test sets for both world models, as these are drawn from the same concept classes as the corresponding training sets. The CNN converges to 100\% accuracy faster than all other architectures; we hypothesise that its architecture allows it to memorise all input-action-output combinations more rapidly than the other networks. The OCL architecture reaches only 0.74 accuracy on both the Translate and Rotate world models, the lowest of all architectures. We report the first 100 epochs here, but we ran OCL for a total of 150 epochs without any change in the results.

On the Translate world model, the APLN achieves perfect accuracy within fewer than 10 epochs, while the APN plateaus at approximately 80\%. On the Distance~1 test set, the Transformer attains roughly 40\% accuracy, whereas the CNN and MLP do not exceed 10\%. OCL achieves 1.3\% on the Translate Distance~1 test set and 0\% on the corresponding Rotate test set. On the Distance~2 test set, the APLN demonstrates full OOD generalisation, again reaching 100\%. The APN performs slightly worse on Distance~2 than on Distance~1, indicating that the axial pointer mechanism without the linear decomposition is insufficient for this world model. The remaining four architectures fail to produce a single correct output image on the Distance~2 test set, indicating that they are unable to learn latent representations supporting OOD generalisation.

All architectures were matched in parameter count for fair comparison (see Table~\ref{table:num_of_params}). This required substantially reducing the dimensionality of the final gMLP layer in the APN (from 256 in the APLN to 32 in the APN), as the APN allocates considerably more parameters to the pixel copying layer. This constraint negatively impacts APN performance. Fig.~\ref{fig:results}(c) presents additional comparisons between the APN and the APLN on the Rotate data sets under relaxed parameter budgets. In two configurations, both networks have approximately 20M parameters (final MLP layer size of 256 for the APN, yielding 20.7M parameters, and 1024 for the APLN, yielding 19.1M). In two further configurations, both networks have approximately 80M parameters (final MLP hidden layer size of 2560 for the APLN and 1024 for the APN). With comparable and substantially larger parameter budgets, the APN exhibits only a marginal improvement over the APLN on the Distance~1 test set. At higher parameter counts, the APN also achieves perfect accuracy on the Distance~0 test set. In the case of OCL, the model already had a comparable number of parameters when used with the hyperparameters suggested by \cite{dedhiaImPromptuIncontext2023}.

As described above, the APLN was designed with an inductive bias towards copying pixels in vertical and horizontal groups. The Rotate world model was specifically constructed to test the importance of such a bias for OOD generalisation. As shown in the Rotate panel (b) of Fig.~\ref{fig:results}, none of the architectures---including the APLN---exhibit any meaningful OOD generalisation on this world model, confirming our hypothesis. Both axial pointer architectures partially solve the Distance~1 test set. The remaining three architectures perform worse than on the Translate world model, with only the Transformer again showing a slight advantage over the CNN and MLP. The Transformer's relative performance across both world models is consistent with the findings of Hupkes et al., who reported that Transformer architectures achieve modestly better OOD generalisation on systematicity tasks compared to other standard architectures.

\subsection{Error Analysis Reveals Non-Compositional Representations}
\label{fails}
Examining the errors each architecture produces on the OOD test sets provides insight into whether a model can at least approximate the correct output. Errors that preserve the correct object shape and/or apply the correct transformation suggest that the network has learned, at least partially, a disentangled representation separating object attributes from the required action.

\begin{figure*}[ht]
\centering
\subfloat[Translate - Distance 1]{%
        \includegraphics[width=0.45\textwidth]{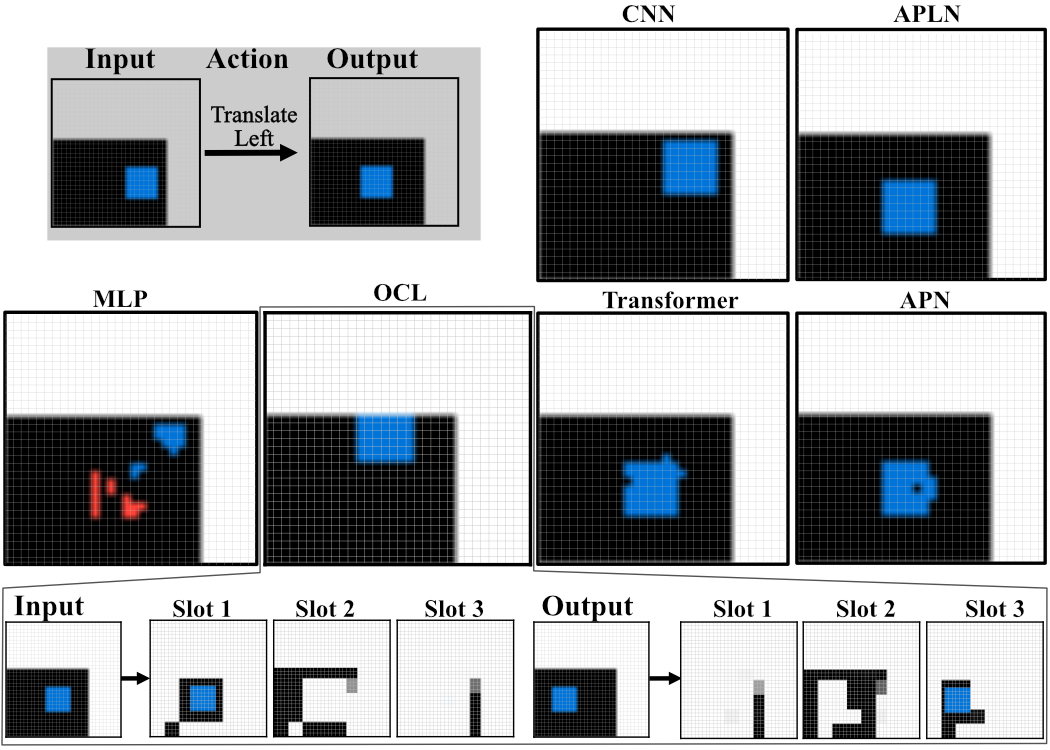}
    }
    \subfloat[Translate - Distance 2]{%
        \includegraphics[width=0.45\textwidth]{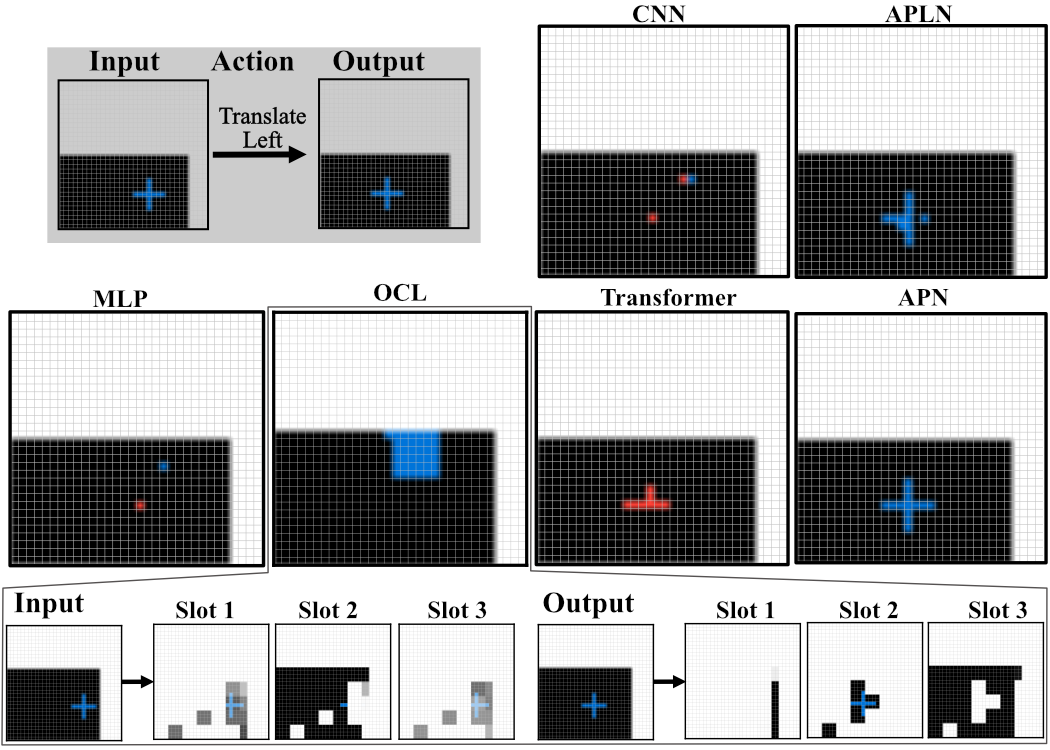}
    }\hfill
    \subfloat[Rotate - Distance 1]{%
        \includegraphics[width=0.45\textwidth]{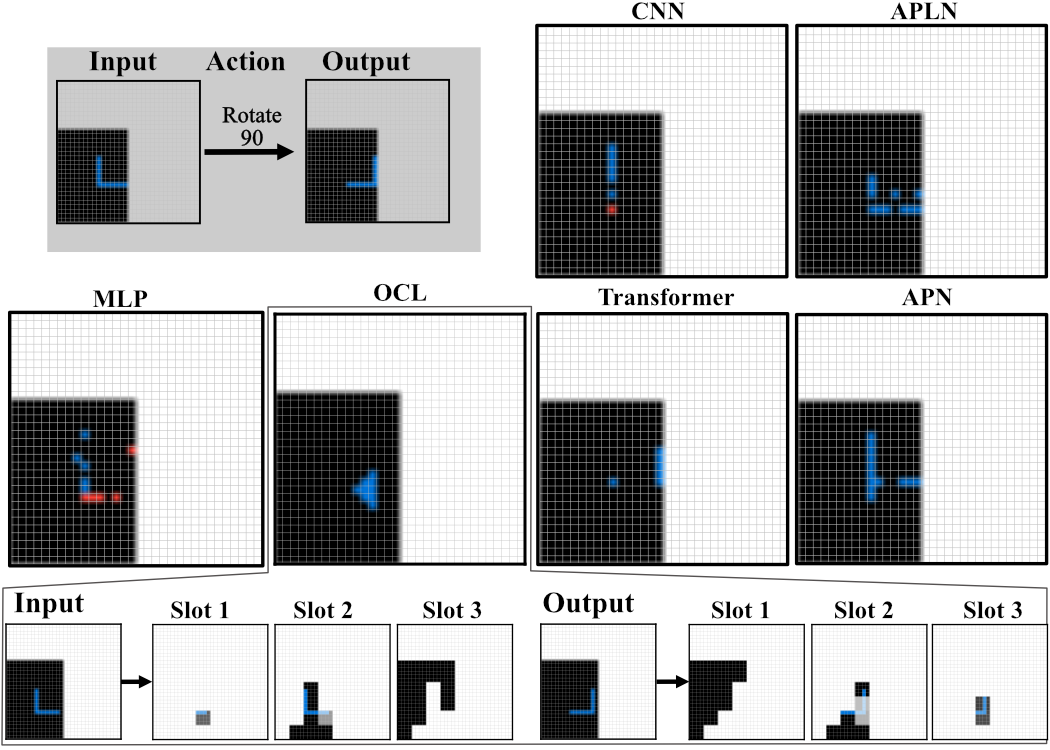}
    }
    \subfloat[Rotate - Distance 2]{%
        \includegraphics[width=0.45\textwidth]{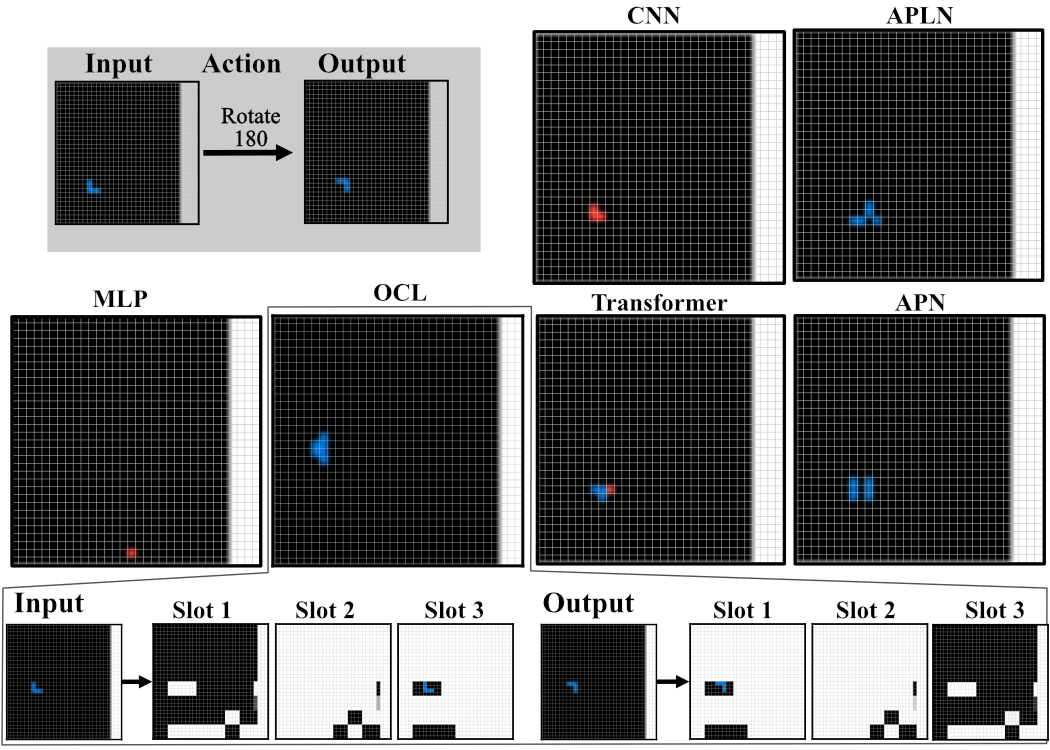}
    }
\caption{Characteristic errors that the networks make in the different test sets. In the top left corner of each sub figure (greyed out) we show the ground truth of the specific datum (input image, action bit and output image). In the Translate - Distance 1 test set (A) the APLN was 100\% correct so we are showing a correct output image. For the OCL error, the input and output images are shown in the third image row of each sub figure. There we also show the three slots discovered by OCL  for each of these input and output images.}
\label{fig:errors}
\end{figure*}

Fig.~\ref{fig:errors} presents characteristic errors across the Distance~1 and Distance~2 test sets for both world models. Across all cases, the MLP produces outputs with the least resemblance to the ground truth, consistent with the known limitations of MLPs on image recognition and generation tasks. CNN errors approximate the correct output only on the Translate Distance~1 test set. In this case, nearly all CNN errors consist of an incorrect translation of the correct object (data not shown), as illustrated in the figure. This suggests that the network successfully copies the object's shape and colour from the input but fails to compose these with the correct action. Instead, it applies only the actions observed during training for each object type, having learned latent features that conflate object appearance with the associated translation. On the remaining test sets (Translate Distance~2 and both Rotate sets), CNN errors exhibit the same lack of grounding to objects and actions as those of the MLP.

\begin{figure*}[h!]
\centering
\subfloat[Pointer network visualisation]{%
        \includegraphics[width=0.95\textwidth]{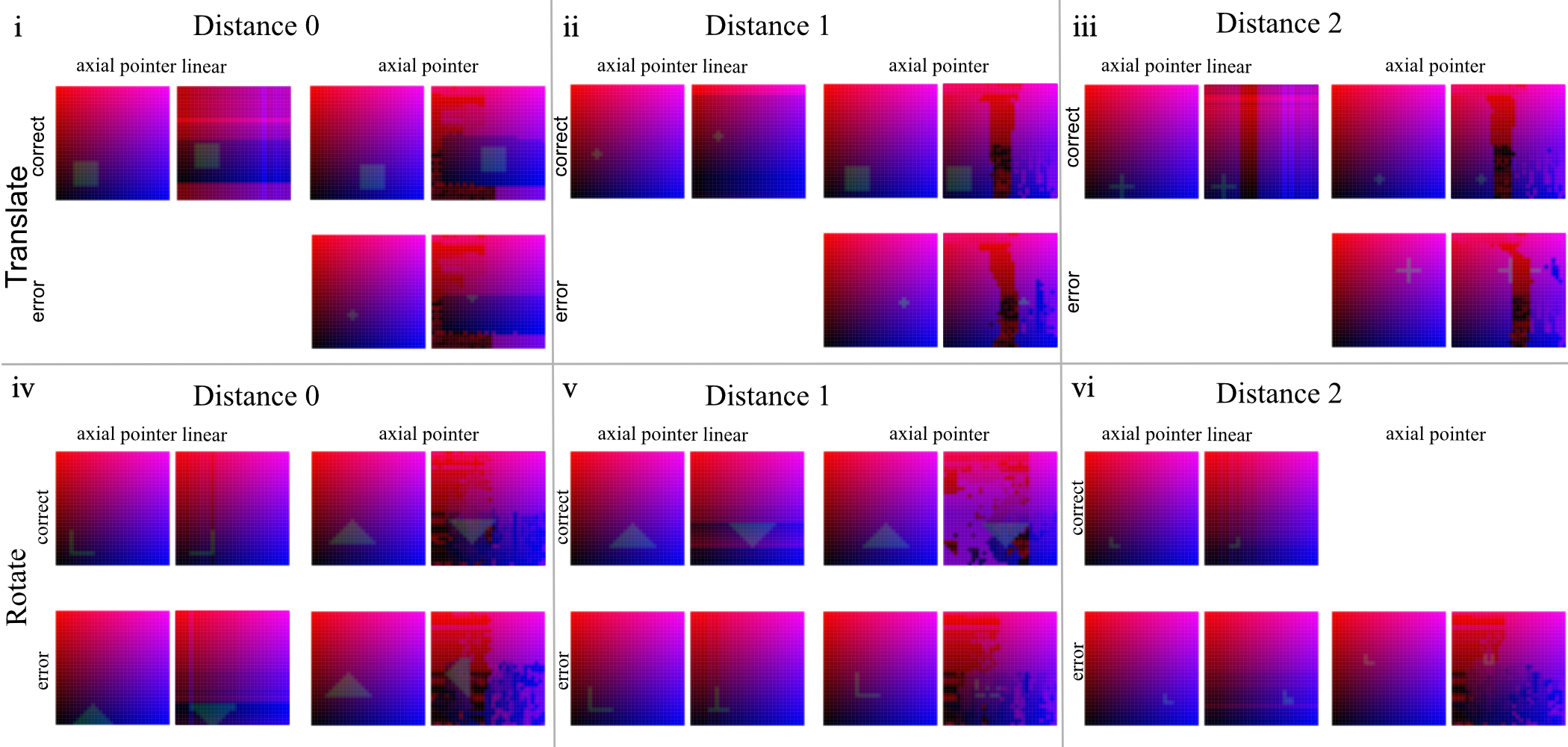}
    }\hfill
    \subfloat[Compositional representation measure]{%
        \includegraphics[width=0.95\textwidth]{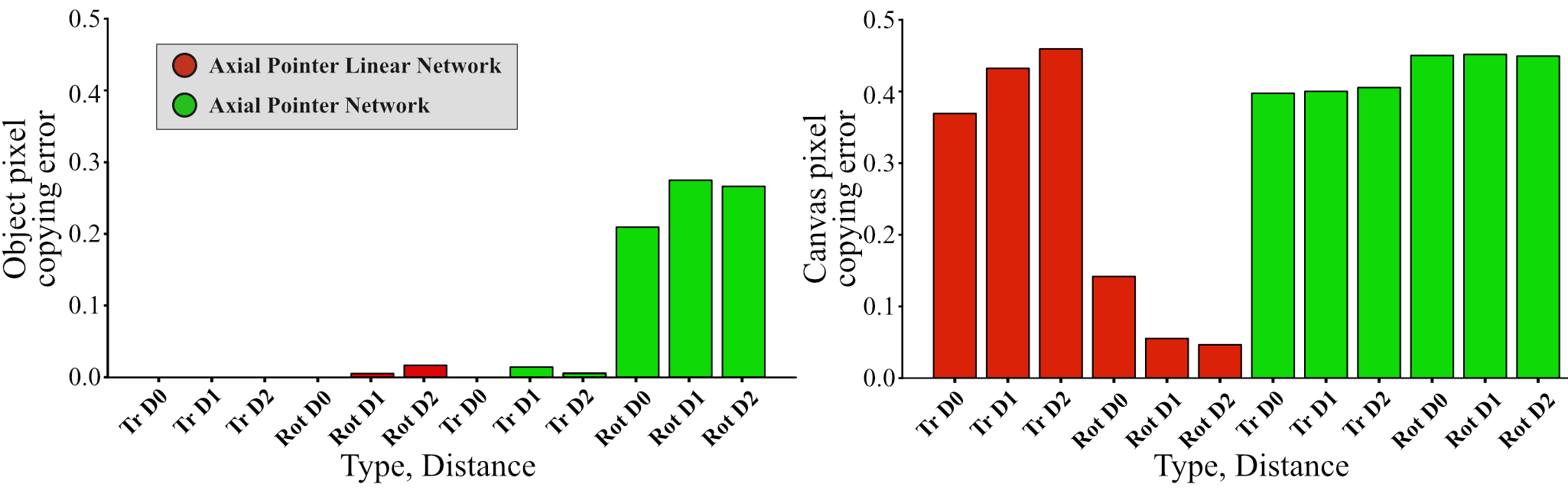}
    }\hfill
\caption{Representation visualisation and compositionality metric. (a) Visualisation of the pointer networks' pixel copying mechanism. The colour coding (black to blue for the x axis and black to red for the y axis) denotes the input index of a pixel. In the input images the colours are distributed in order. In the output images the colours denote the pixel index of the input image each pixel was copied from. The light green tinge on some pixels in the images show the pixels that were part of the coloured objects in the input. For each test set (world model + distance) we show how each of the two axial pointer networks copies pixels from input to output both in the case of correct prediction (lines of correct) and of a wrong one (lines of error). For each network, world model, distance combination the left image is the input and the right the output image. Where we provide no visualisations (e.g. APLN, Translate, Distance 0, error or APN, Rotate, Distance 2, correct) is because for these cases there were no samples generated. (b) The results of the measure of compositional representation M for all both axial pointer networks (APLN: red bars, APN: green bars) for all world models and distances. The left bar plot shows the object pixel copying error while the right one shows the error in the copying of the rest of the pixels (canvas pixels).}
\label{fig:visualisation}
\end{figure*}

The Transformer consistently produces errors closer to the ground truth across all test sets, generating approximately correct objects (with minor pixel additions or omissions) and applying approximately correct transformations. On the Distance~2 sets, erroneous pixels may also have incorrect colour values, a phenomenon not observed at Distance~1 (data not shown). These error characteristics are again consistent with the findings of Hupkes et al.\ regarding the Transformer's superior OOD performance.

The majority of the OCL network's mistakes are either in the rule (creating the correct object but not positioning it appropriately) or cases where both the object and the rule are wrong. In a large proportion of the mistaken-object cases (both with and without the correct rule), the network generates one of the other possible objects. For example, in the Rotate/Distance~2 case, instead of a blue angle rotating 180\degree{}, the model often generates a 90\degree{}-rotated small blue pyramid (data not shown, but see Fig.~\ref{fig:errors}(c)(OCL) for an example). In the remaining cases where the object is wrong, the network misses the correct object by a few pixels (e.g., a truncated pyramid---see Fig.~\ref{fig:errors}(d)(OCL)---or a square with a few missing or extra pixels). Table~\ref{table:ocl_errors_per_dist} gives, for each world model and distance, the percentage of mistakes across four error types: (i) correct shape and rule (final positioning) but wrong colour, (ii) correct rule but wrong shape, (iii) correct shape but wrong rule, and (iv) both shape and rule wrong (for types ii, iii, and iv we ignore colour). These results emphasise that, despite OCL's bias towards object recognition, the majority of mistakes still involve a combination of misshaped objects and mistaken rule application.

Despite the OCL architecture's focus on object recognition, object segmentation, and compositional rule application, the model fails on our world models notwithstanding their compositional simplicity. We propose that our world models, combined with the discrete, full-correct-image metric they allow, are a more appropriate test not just of compositionality but of actual OOD generalisation. As shown by the errors in Fig.~\ref{fig:errors}, the OCL architecture manages to assign only roughly the correct object to the input slots (with a large number of extra pixels and, sometimes, some pixels missing). Despite this, the output generated by the network is, most of the time, a combination of the wrong object and the wrong rule application. The objects finally generated are sometimes objects the network has seen in other samples, and sometimes mangled objects with an obvious resemblance to objects seen in the training set. Our initial hypothesis for this failure mode is that the network is designed to compose over objects and object attributes (assigned to slots) but its attention over slots composition layer is unable to deal with composition over actions (because the slots do not encode the action space of the world models). It is possible that increasing the number of slots to match the number of possible position/orientation combinations for each object (thus encoding all possible object transformations in this large set of slots) might allow the network to produce the correct object output. Were this confirmed, it would suggest that any resulting improvement stems from memorising the training set rather than from genuine OOD-capable generalisation.

The structure of the training and testing data sets implies an imbalance that could affect the way the networks learn. In the training set, there are three times as many samples of one rule (\textbf{Up} and \textbf{90}) as of the other. This could bias the networks towards one rule over the other. To better understand the impact of this imbalance, we show in Table~\ref{table:ocl_errors_per_rule} the number of samples across error types for the four rules (\textbf{Up}, \textbf{Left}, \textbf{90}, and \textbf{180}). It is evident that this 3:1 imbalance affects the number of fully correct and fully wrong samples depending on the rule. The number of fully correct samples for the \textbf{Up} and \textbf{90} rules is about three times that for the \textbf{Left} and \textbf{180} rules (e.g., 532 correct for \textbf{Up} vs.\ 247 for \textbf{Left}), while the opposite ratio holds for the all-wrong sample category (e.g., 1321 for \textbf{Left} vs.\ 408 for \textbf{Up}). This transfer of the sample-distribution imbalance into the distribution of OCL error types underscores that even networks with object-based inductive biases, presumed to solve their tasks through object and rule composition, can fail to learn in a genuinely OOD-generalisable way and instead remain in the regime of learning irrelevant statistical correlations.

The axial pointer networks produce substantially more accurate outputs on the Translate world model, where errors typically consist of small numbers of omitted or spurious pixels. This is not the case on the Rotate world model, where the pixel copying operates in a near-random fashion, though concentrated in the spatial region where the correct object should appear. As expected, given that the APN and APLN operate by copying pixels from input to output, these architectures never produce pixels of incorrect colour. Overall, the error patterns across all architectures indicate that even in cases where the aggregate error rate is low but non-zero, none of the networks have learned genuinely compositional representations of either the objects or the required transformations.

\begin{table}[h!]
\begin{tabular}{ |p{2.5cm}||p{0.5cm}|p{0.5cm}|p{0.5cm}|p{0.5cm}|p{0.5cm}|p{0.5cm}|}
 \hline
\textbf{Error type} & \textbf{T/D0}  & \textbf{T/D1}  & \textbf{T/D2}& \textbf{R/D0}  & \textbf{R/D1}  & \textbf{R/D2}\\
  \hline
Fully correct. & 0.77& 0.01& 0 & 0.74& 0 & 0  \\
Correct shape and rule. Wrong colour. & 0 & 0.02& 0 & 0 & 0.01& 0 \\
Correct rule. Wrong shape. & 0.08& 0.12& 0.05& 0.08& 0.53& 0.17\\
Correct shape. Wrong rule. & 0.03& 0.18& 0.02& 0.02& 0.01& 0 \\
All wrong. & 0.12& 0.67& 0.93& 0.16& 0.46& 0.83\\
\hline
\end{tabular}
\\
\caption{Percentage of types of errors for each world model / distance for the OCL architecture.}
\label{table:ocl_errors_per_dist}
\end{table}

\begin{table}[h!]
\begin{tabular}{ |p{1.0cm}||p{0.9cm}|p{0.9cm}|p{0.9cm}||p{0.9cm}|p{0.9cm}|}
 \hline
\textbf{Rule} & \textbf{Correct}  & \textbf{Wrong colour}& \textbf{Wrong shape}  & \textbf{Wrong rule}  & \textbf{All wrong}\\
  \hline
\textbf{Left} & 247&17& 137 & 200& 1321\\
\textbf{Up} & 532 & 0 & 107 & 28 & 408\\
\textbf{90} & 511 & 11 & 104 & 16 & 398\\
\textbf{180} & 225 & 0 & 684 & 4 & 1044\\
\hline
\end{tabular}
\\
\caption{Number of types of errors for each rule for the OCL architecture.}
\label{table:ocl_errors_per_rule}
\end{table}

\subsection{The Copying Layer of the Axial Pointer Architectures Reveals Non-Compositional Structure}
\label{copy_layer}
As noted above, the axial pointer architecture enables direct visualisation (Fig.~\ref{fig:visualisation}(a)) of the pixel-level copying operation in the final layer, through which we can assess whether any network has learned to organise its copying operations in a compositionally structured manner.

The first observation is that the APLN groups copied pixels into rows and columns, as intended by design. Comparing the APLN's copying patterns between the Translate and Rotate world models, it is precisely this row--column grouping that enables strong OOD performance on the Translate model while causing failure on the Rotate model. Figs.~\ref{fig:visualisation}A-iv, A-v, and A-vi show that, on correct predictions in the Rotate model, the APLN occasionally leverages its bias to approximate the rotated output by rearranging multiple rows and columns.

The visualisations for the APN across all world model and distance combinations reveal that this architecture also copies pixels in spatially coherent groups. However, these groups exhibit noisier boundaries and less consistent internal structure compared to the row--column groups of the APLN. Despite having the capacity to form pixel groups corresponding to objects and to learn representations of their transformations as affine operations (translation and rotation), the APN never develops such latent features.

To provide a quantitative assessment of the networks' failure to form latent variables corresponding to the compositional structure of the tasks, we define a metric of divergence from the compositional representation. This metric measures the average discrepancy between the observed pixel displacement and the displacement expected under a compositionally correct operation that acts solely on the coloured objects. Specifically, the metric $M$ is defined as:
\[M = \left[\frac{\frac{1}{C}\sum_{c=0}^{C}\lVert p_c^* - p_c\rVert^2 }{ \frac{1}{C+B}\sum_{i=0}^{C+B}\lVert p_i^{rand} - p_i\rVert^2},\; \frac{\frac{1}{B}\sum_{b=0}^{B}\lVert p_b^* - p_b\rVert^2 }{ \frac{1}{C+B}\sum_{i=0}^{C+B}\lVert p_i^{rand} - p_i\rVert^2}\right] \]
where $C$ denotes the number of coloured (object) pixels and $B$ the number of black (canvas) pixels in the image. The term $p_c^*$ is the expected displacement of object pixel $c$ under compositionally correct behaviour, i.e., the displacement that would result from detecting and transforming the coloured object. The term $p_b^*$ is the expected displacement of canvas pixel $b$, defined as~0. The terms $p_c$ and $p_b$ are the actual displacements applied by the network to object pixel $c$ and canvas pixel $b$, respectively. The term $p_i^{rand}$ is the displacement of pixel $i$ under a random permutation of all pixel positions.

The metric decomposes into two components. The first quantifies how accurately the network displaces object pixels according to the task's prescribed action. The second measures the degree to which the network distinguishes between object and canvas pixels, thereby assessing whether the copying layer has acquired the concept of a coloured shape. Fig.~\ref{fig:visualisation}(b) shows that in most network/distance combinations, the networks succeed in detecting and correctly displacing the object pixels (with the exception of the APN on the Rotate world model). However, in all cases they fail to discriminate between object and canvas pixels, with most configurations exhibiting approximately 50\% divergence from the random baseline for canvas pixel displacement. A notable exception is the APLN at Distance~1 and Distance~2 on the Rotate world model, where the network successfully minimises canvas pixel displacement but, due to its inductive bias, still displaces entire rows or columns (as shown in Figs.~\ref{fig:visualisation}A-v and A-vi).

In order to validate the metric we run it on the results of a hand crafted function that took the input images and rule denoting language bits as input, discovered the input objects and followed the correct rule to reposition the object, creating the output image. At the same time it kept track of each pixel displacement (much like the APN and APLN architectures). Running the metric on these displacements we generated the metric's theoretical minimum for the object and canvas pixels for all world models and distances (see Table~\ref{table:metric_validation}).

\begin{table}[h!]
\begin{tabular}{ |p{2.5cm}||p{0.4cm}|p{0.4cm}|p{0.4cm}|p{0.6cm}|p{0.6cm}|p{0.6cm}|}
 \hline
\textbf{Pixel type} & \textbf{T/D0}  & \textbf{T/D1}  & \textbf{T/D2}& \textbf{R/D0}  & \textbf{R/D1}  & \textbf{R/D2}\\
  \hline
Mean Object pixel displacement. & 0 & 0 & 0 & 0.0036 & 0.0049 & 0.005  \\
Mean Canvas pixel displacement. & 0 & 0 & 0 & 0 & 0 & 0 \\
\hline
\end{tabular}
\\
\caption{Object and Canvas mean pixel displacements for the results of a function crafted to produce the correct output.}
\label{table:metric_validation}
\end{table}

\section{Discussion}
\label{discussion}
Our experiments reveal a fundamental gap between OOD task performance and the acquisition of compositional representations. Both the three baseline networks (MLP, CNN, and Transformer) and the composition-biased OCL model achieve very poor results in the Distance~1 and Distance~2 settings for both world models. While the APLN achieves near-perfect accuracy on the Translate world model, inspection of its learned representations demonstrates that this success is attributable to a hand-crafted inductive bias rather than to the emergence of genuinely compositional latent structure. These findings have direct implications for how the community evaluates claims of compositional generalisation: benchmark accuracy on held-out distribution shifts, even when carefully controlled, does not constitute sufficient evidence that a model has learned reusable, composable primitives.

Several limitations of the present study should be acknowledged. First, we evaluate a relatively small set of architectures. Extending the analysis to a broader range of models---including graph neural networks, state-space models, and neurosymbolic architectures specifically designed for systematic generalisation---would provide a more comprehensive picture of the current landscape. The OCL model used here was the only example of such an object-based, composition-capable architecture, and its failure to generalise OOD highlights the need for more extensive testing of such models. Second, hyperparameter search was limited in scope. Because all models (except OCL) achieve ceiling performance on the in-distribution test set, there is no in-distribution validation signal to drive automated hyperparameter optimisation for OOD performance. We consider it unlikely that alternative hyperparameter configurations would qualitatively alter the relative ordering of the tested architectures on OOD sets, though this remains an open empirical question. For OCL, we kept the hyperparameters practically identical to those used by its creators to demonstrate its compositional capabilities. We believe this yields a fairer comparison than tuning the hyperparameters ourselves to reach ceiling performance on the training set. Third, the current world models employ atomic actions (single translations or rotations). We do not explore settings with compositional action spaces---i.e., tasks where the target transformation is itself a composition of primitive operations---which would enable evaluation of a broader class of models, including language-conditioned architectures such as that of Lake and Baroni. That said, the OCL example points to the need to distinguish between complex compositional chains over object attributes, where a large number of bespoke networks achieve good results, and OOD generalisation over both object and rule composition, where, we argue, the same architectures may be less appropriate. Keeping the world models compositionally simple but demanding in terms of OOD capability is one way to partially achieve this.

Several directions naturally extend the present framework. A high-priority extension is the introduction of compositional action spaces, where output transformations are sequences or combinations of primitive operations. This would enable direct comparison with meta-learning and in-context learning approaches that rely on linguistic or symbolic task specifications. Additionally, scaling the world models to higher-dimensional observation spaces and richer object ontologies would test whether the dissociation between performance and compositional representation persists under more complex distributional shifts.

On the evaluation side, the development of automated, architecture-agnostic probing methods for compositional structure in latent spaces---beyond the pointer-network visualisation employed here---would make the proposed evaluation paradigm applicable to arbitrary deep networks. Techniques from mechanistic interpretability, such as activation patching and causal tracing, offer promising tools for this purpose.

Finally, it would be valuable to investigate whether training objectives that explicitly incentivise compressed, factored representations---such as information-bottleneck regularisation or disentanglement penalties---can bridge the gap between OOD performance and genuine compositionality that our results expose. Connecting these findings to the complexity-theoretic framework of Elmoznino et al.\ through empirical measurement of representational Kolmogorov complexity would help ground future compositionality claims in more rigorous theoretical criteria.

\section{Acknowledgments}
GD and SS acknowledge the support of the Engineering and Physical Sciences Research Council (EPSRC) grant ``UKRI1076 Element-compositor methods for out-of-distribution machine learning''.\\
For the purpose of open access, the authors have applied a
Creative Commons Attribution (CC BY) licence to any Author
Accepted Manuscript version arising.\\
We used Claude Opus 4.6 Extended as an editor. The whole
manuscript was written manually by the authors and then given
to Claude to do small changes in the format of phrases and
the choice of certain words. The proposed changes were then
re-reviewed by the authors.

\bibliographystyle{IEEEtran}
\bibliography{V2/bibliography/bibliography}
\vspace{1.0cm}
\textbf{George Dimitriadis} obtained a B.S. in Biology from the Aristoteles University of Thessaloniki, Greece, a B.S. in Physics and a B.S. in Psychology from the Open University, UK and a Ph.D. in Protein Engineering from the Department of Physics and Astronomy of the University of Leeds, UK. He was a Research Associate in Experimental Neuroscience at the Donders Centre for Cognition in Radboud University, the Netherlands and a Senior Research Associate at the Sainsbury Wellcome Centre (SWC), UCL, UK. He was a Joint Senior Research Fellow in Experimental and Theoretical Neuroscience in SWC and the Gatsby Computational Neuroscience Unit, UCL, UK. He is currently a Research Associate in the School of Electronic Engineering and Computer Science in the University of Essex, UK.     
\\
\\
\textbf{Spyridon Samothrakis} received the Ph.D. degree in computer science in 2014, with a focus on optimal game playing in multi-player games. He is currently a Senior Chief Scientific Adviser and Senior Lecturer at the School of Computer Science and Electronic Engineering (CSEE), University of Essex.  His current research interests include meta-learning and reinforcement learning.

\end{document}